\PassOptionsToPackage{square,numbers,sort&compress}{natbib}
\documentclass{article}

\usepackage[final,sglblindworkshop]{neurips_2025}
\workshoptitle{Evaluating the Evolving LLM Lifecycle}

\usepackage[utf8]{inputenc} % allow utf-8 input
\usepackage[T1]{fontenc}    % use 8-bit T1 fonts
\usepackage{hyperref}       % hyperlinks
\usepackage{url}            % simple URL typesetting
\usepackage{booktabs}       % professional-quality tables
\usepackage{amsfonts}       % blackboard math symbols
\usepackage{nicefrac}       % compact symbols for 1/2, etc.
\usepackage{microtype}      % microtypography
\usepackage{xcolor}         % colors
\usepackage{amsmath}
\usepackage{graphicx}
\usepackage{subcaption}
\usepackage{multirow}
\usepackage{makecell}   % for \makecell
\usepackage{arydshln}
\usepackage[whole]{bxcjkjatype}
\usepackage{pifont}
\usepackage{enumitem}
\usepackage{cleveref}
\usepackage{subcaption}

\title{\textit{PULSE}: Practical Evaluation Scenarios \\for Large Multimodal Model Unlearning}

\author{%
Tatsuki Kawakami\textsuperscript{1} \quad
Kazuki Egashira\textsuperscript{1} \quad
Atsuyuki Miyai\textsuperscript{1} \quad
Go Irie\textsuperscript{2} \quad
Kiyoharu Aizawa\textsuperscript{1} \textsuperscript{2} \\
[1mm]
\textsuperscript{1}The University of Tokyo \quad
\textsuperscript{2}Tokyo University of Science \\
\texttt{kawakami@hal.t.u-tokyo.ac.jp} \\
\vspace{-0.8em}
}

\begin{document}

\maketitle

\begin{abstract}
  In recent years, unlearning techniques, which are methods for inducing a model to ``forget'' previously learned information, have attracted attention as a way to address privacy and copyright concerns in large language models (LLMs) and large multimodal models (LMMs). While several unlearning benchmarks have been established for LLMs, a practical evaluation framework for unlearning in LMMs has been less explored. 
Specifically, existing unlearning benchmark for LMMs considers only scenarios in which the model is required to unlearn fine-tuned knowledge through a single unlearning operation. 
In this study, we introduce \textbf{PULSE} protocol for realistic unlearning scenarios for LMMs by introducing two critical perspectives: (i) \textit{\textbf{P}re-trained knowledge \textbf{U}nlearning} for analyzing the effect across different knowledge acquisition phases and (ii) \textit{\textbf{L}ong-term \textbf{S}ustainability \textbf{E}valuation} to address sequential requests.
  We then evaluate existing unlearning methods along these dimensions. Our results reveal that, although some techniques can successfully unlearn knowledge acquired through fine-tuning, they struggle to eliminate information learned during pre-training. Moreover, methods that effectively unlearn a batch of target data in a single operation exhibit substantial performance degradation when the same data are split and unlearned sequentially.
\end{abstract}

\section{Introduction}

In recent years, Large Language Models (LLMs)~\cite{gpt3} and Large Multimodal Models (LMMs)~\cite{lmm_survey} have achieved great success across a variety of tasks. However, because their training data can include personal information and copyrighted content, concerns have been raised about privacy and intellectual property infringement. Against this backdrop, there has been growing interest in (approximate) unlearning, which is a machine learning technique that aims to preserve performance on designated retention tasks while degrading performance on unlearning tasks~\cite{mu_for_linear, mu_for_ramdomtree}. Recently, methods tailored specifically for LLMs and LMMs have also been proposed~\cite{ga, whos_hp, siu}.

As interest in unlearning research for LLMs and LMMs grows, there is an increasing need to develop a unified evaluation methodology for these techniques~\cite{tofu, muse}. However, no practical and comprehensive evaluation framework currently exists for unlearning in LMMs. As prior work, MLLMU-Bench~\cite{mllmu_bench} provides a benchmark for LMM unlearning, but from a practical standpoint it is insufficient because
(1) it only considers unlearning on the data used in the most recent fine-tuning, casting doubt on whether it can be extended to the pre-trained knowledge, and (2) it only accounts for the scenario where unlearning request occurs only once, necessitating evaluation over sequential unlearning operations.

\begin{figure*}[t]
    \centering
    \begin{subfigure}{1\textwidth}
        \includegraphics[width=\textwidth]{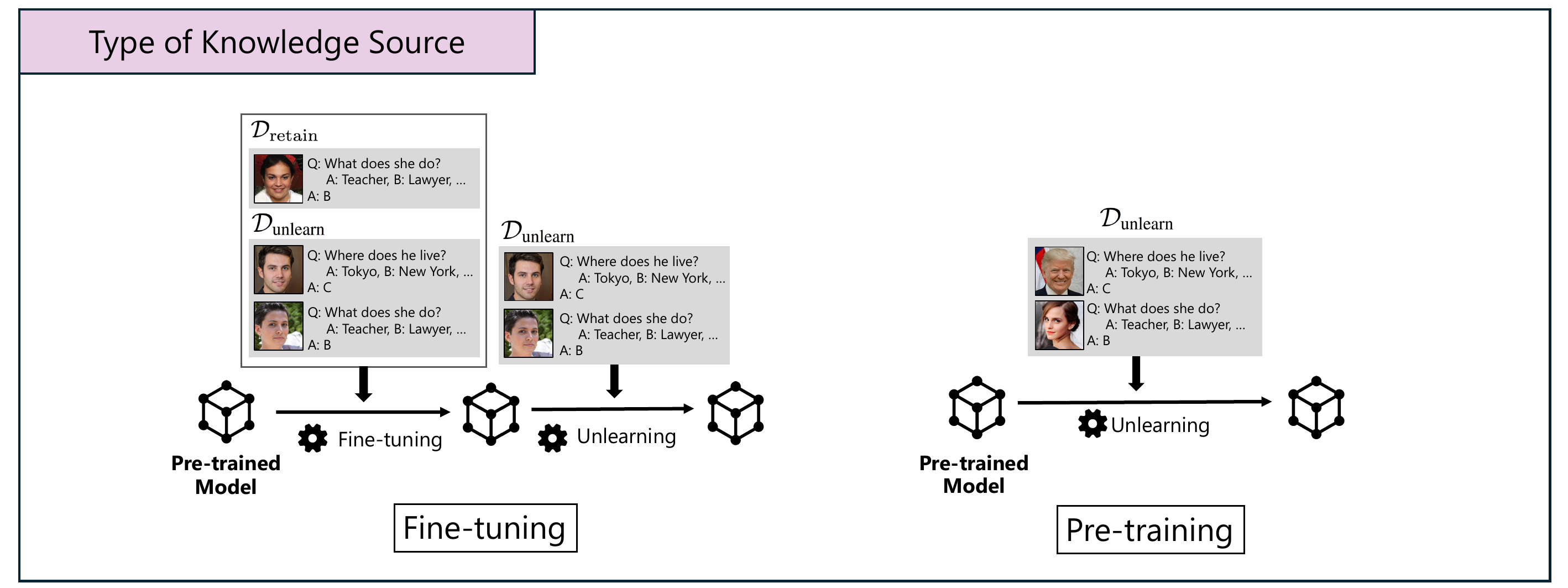}
        \caption{\textbf{Type of Knowledge Source:} (left): As in prior work~\cite{mllmu_bench}, we first fine-tune the model and assess unlearning of the targeted samples within the fine-tuned dataset. (right): We additionally evaluate whether existing methods can unlearn the knowledge that is obtained in the pre-trained phase.}
        \label{fig:teasers1}
    \end{subfigure}
    \hfill
    \begin{subfigure}{1\textwidth}
        \includegraphics[width=\textwidth]{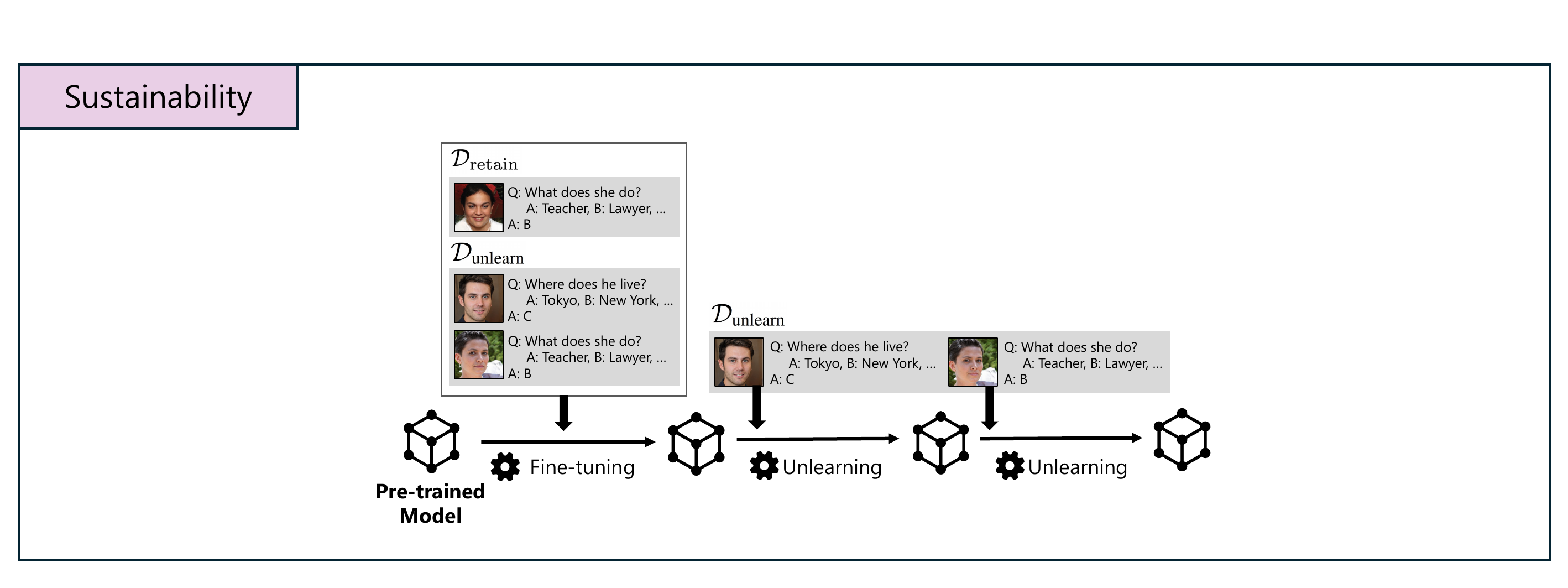}
        \caption{\textbf{Sustainability:} We split the unlearning target $\mathcal{D}_{\text{unlearn}}$ into a few subset and perform unlearning sequentially.}
        \label{fig:teasers2}
    \end{subfigure}
    \caption{\textbf{Our PULSE Pipelines.}}
    \label{fig:teasers}
\end{figure*}

\paragraph{Our Work: A practical Evaluation Framework for Unlearning in LMMs}
In this study, we introduce \textbf{PULSE} protocol, which is a new evaluation protocol that addresses two key settings:  (1) \textit{\textbf{P}re-trained knowledge \textbf{U}nlearning}: unlearning knowledge obtained during pre-training and (2) \textit{\textbf{L}ong-term \textbf{S}ustainability \textbf{E}valuation}: addressing multiple sequential unlearning requests. Figure~\ref{fig:teasers} shows our PULSE pipelines of each experiment. Table~\ref{tab:bench_comparison} compares our evaluation method with existing benchmarks. By using our protocol to assess existing unlearning approaches, we aim to establish a more practical evaluation foundation for LMM unlearning.

Experiments based on our proposed evaluation protocol revealed that existing unlearning methods fail to deliver adequate performance in (1) scenarios requiring the unlearning of pre-trained knowledge and (2) scenarios necessitating multiple, sequential unlearning operations. These findings provide important insights for the future design of unlearning techniques for LMMs.

\paragraph{Contribution}
\begin{itemize}[itemsep=0.1em, topsep=0.1em]
\item We propose \textbf{PULSE} protocol, designed to evaluate (i) \textbf{P}re-trained knowledge \textbf{U}nlearning and (ii) \textbf{L}ong-term \textbf{S}ustainability \textbf{E}valuation in large multimodal models (LMMs).
\item Through our \textbf{PULSE} protocol, we show that existing unlearning techniques are ineffective when the target is knowledge acquired during pre-training, even though they perform well for fine-tuned knowledge.
\item We also find that model performance degrades significantly when subjected to multiple sequential unlearning requests, indicating that current approaches remain impractical for real-world deployment.
\end{itemize}

\section{Related Work}

\begin{table}[t]
    \caption{\textbf{Comparison with Prior Evaluation Methods.} We assess not only the unlearning of fine-tuned knowledge in a single request, but also (i) pre-trained knowledge unlearning and (ii) sustainable unlearning against multiple unlearning requests, providing the first comprehensive evaluation protocol for unlearning in LMMs.}
    \label{tab:bench_comparison}
    \centering
    \setlength{\tabcolsep}{3pt}
    \resizebox{1\linewidth}{!}{
    \begin{tabular}{lcccc}
    \toprule
        & \textbf{Target Model} 
        & \makecell[c]{\textbf{Unlearning of}\\\textbf{Fine-Tuned Knowledge}} 
        & \makecell[c]{\textbf{Unlearning of}\\\textbf{Pre-trained Knowledge}} 
        & \textbf{Sustainability} \\
    \midrule
    \textbf{MUSE~\cite{muse}}            & LLM & \ding{51} &              & \ding{51} \\
    \textbf{TOFU~\cite{tofu}}            & LLM & \ding{51} &              &           \\
    \textbf{\citet{yao2024machine}}      & LLM & \ding{51} & \ding{51}    &           \\
   \cdashline{1-5}
    \textbf{MLLMU-Bench~\cite{mllmu_bench}}     & LMM & \ding{51} &              &              \\
     \textbf{PULSE(Ours)}            & LMM & \ding{51} & \ding{51} & \ding{51} \\
    \bottomrule
    \end{tabular}
    }
\end{table}

\subsection{Methodology of Unlearning}

As one unlearning method for neural networks, Gradient Ascent (GA) has been proposed. Improved variants such as GA with added regularization~\cite{ga} and NPO~\cite{npo} have also been introduced. These techniques are all widely used~\cite{tofu,muse,siu,mllmu_bench} and have been shown to work to some extent on LLMs and LMMs~\cite{muse,mllmu_bench}. However, whether they offer sufficient unlearning effectiveness for pre-trained knowledge and sustainability in the context of LMM unlearning remains unverified. This study aims to incorporate these perspectives and evaluate their performance in realistic use cases.

In addition, SIU~\cite{siu} has been proposed as an LMM-specific unlearning method, but SIU is limited to multimodal tasks and does not address unlearning for text-only tasks. As we argue in Section~\ref{subsec:setup}, we believe that ensuring that ``no information about the unlearning targets is leaked regardless of the task'' is crucial; therefore, SIU is not included among the methods evaluated in this study.  

\subsection{Benchmarks for Unlearning}

MUSE~\cite{muse}, a benchmark for unlearning in LLMs, evaluates effectiveness and generality from various perspectives. Notably, it employs metrics that consider practical application aspects such as ``sustainability,'' which reflects the ability to handle continuous unlearning requests. We believe that sustainability is also important for unlearning in LMMs, and have adopted it as an evaluation criterion in this study.

For unlearning in LMMs, MLLMU-Bench~\cite{mllmu_bench} was proposed as one of the earliest benchmarks. It uses a dataset of 500 fictional individuals for experimental evaluation. In our work, we also utilize the publicly available dataset from this benchmark.
While MLLMU-Bench provides a foundation for evaluating unlearning in LMMs, it places less emphasis on the evaluation of unlearning pre-trained knowledge and of sustainability.
Therefore, in addition to the fine-tuned knowledge unlearning (left side of Figure~\ref{fig:teasers1}, we evaluate unlearning of pre-trained knowledge (right side of Figure~\ref{fig:teasers1}) and sustainability (Figure~\ref{fig:teasers2}), evaluating existing unlearning methods from a viewpoint grounded in realistic use cases.

\section{PULSE Protocol}
In addition to the typical unlearning evaluation pipeline that first fine-tunes the unlearning target samples and then conduct unlearning on it, our PULSE evaluates (i) unlearning on pre-trained knowledge and (ii) sustainability against multiple unlearning requests.
In this section, we introduce the general problem setup (\Cref{subsec:setup}) and the detailed setup of evaluation on each perspective (\Cref{subsec:setup_finetune,subsec:setup_pretrain,subsec:setup_sustainability}).

\subsection{Problem Formulation}
\label{subsec:setup}

Let $\mathcal{D}_{\text{unlearn}}$ denote the data to be unlearned and $\mathcal{D}_{\text{retain}}$ the data to be retained. 
In general, the evaluation metrics for unlearning methods consist of two aspects: the unlearning performance on the unlearning target, $\mathcal{D}_{\text{unlearn}}$ (\textbf{effectiveness}) and the accuracy retention on the irrelevant data $\mathcal{D}_{\text{retain}}$ (\textbf{generality})~\cite{tofu}.
Effectiveness and generality exist in a trade-off relationship. For example, attempting to fully unlearn an individual may inadvertently remove knowledge about other individuals or concepts; conversely, preserving full generality may degrade 
unlearning effectiveness. Therefore, effectiveness and generality must always be evaluated simultaneously.

As a realistic use case for unlearning in LMMs, consider forgetting information about a specific individual. In such a scenario, it is desirable that the model does not output any information about the target person, regardless of whether an image of that person is provided as input. Therefore, in this study we conduct experiments under a setting in which the model must not reveal any information about the unlearning target in both multimodal tasks and text-only tasks (Figure~\ref{fig:task}).

\subsection{Fine-tuned Knowledge Unlearning}
\label{subsec:setup_finetune}
The left side of Figure~\ref{fig:teasers1} illustrates the pipeline used in our fine-tuned knowledge unlearning experiment. Following standard practices from benchmarks in both LLMs~\cite{tofu, muse} and LMMs~\cite{mllmu_bench}, a subset of the fine-tuning knowledge, denoted as $\mathcal{D}_{\text{unlearn}}$, is selected as the unlearning target. The model then unlearns this subset in a single operation.

\subsection{Pre-trained Knowledge Unlearning}
\label{subsec:setup_pretrain}

Existing unlearning benchmarks such as TOFU~\cite{tofu}, MUSE~\cite{muse}, and MLLMU-Bench~\cite{mllmu_bench} evaluate only the unlearning of knowledge gained via fine-tuning, but real-world use cases may require forgetting knowledge obtained during pre-training. Moreover, the technical difficulty of unlearning pre-trained knowledge may differ significantly from that of fine-tuned knowledge. Therefore, we evaluate performance specifically on the unlearning of pre-trained knowledge.

The right side of Figure~\ref{fig:teasers1} shows the pipeline for the pre-trained knowledge unlearning experiment. Here, the knowledge obtained during pre-training is treated as $\mathcal{D}_{\text{unlearn}}$ and unlearned in a single unlearning step. To ensure the pre-trained model initially possesses sufficient knowledge of the target individuals, we picked up individuals on which a pre-trained model performs well from existing dataset of celebrities.

While recent work has explored pre-trained knowledge unlearning in LLMs~\cite{yao2024machine}, our approach differs in key ways. Specifically, their method samples $\mathcal{D}_{\text{unlearn}}$ and $\mathcal{D}_{\text{retain}}$ directly from the pre-training data, which requires access to that data. In contrast, our approach is based on the model's actual behavior (i.e., identifying individuals the model demonstrably ``knows''). Although this does not formally guarantee the individual's inclusion in the pre-training data, such inclusion is highly likely and, more importantly, this strategy is more practical when pre-trained corpus is not fully disclosed.

\subsection{Sustainability}
\label{subsec:setup_sustainability}

In real-world unlearning scenarios, data owners request the model creators to prevent the model from outputting information about their data, and with each such request, the model creator must apply unlearning to the existing model. Consequently, cases of multiple unlearning operations on a single model occur frequently. Therefore, a practical unlearning method must maintain performance even after repeated unlearning, and we evaluate this capability in our study.

The importance of such evaluation axis is first proposed by MUSE~\cite{muse} for LLMs; here, we extend it to the multimodal setting for LMMs. Figure~\ref{fig:teasers2} illustrates the pipeline of sustainability evaluation. In contrast to the single-step unlearning in~\Cref{subsec:setup_finetune}, here we divide $\mathcal{D}_{\text{unlearn}}$ into several subsets and make the target model unlearn them sequentially. During this process, we track how the model's generality and effectiveness change after each operation.

\begin{figure}[t]
\centering
    \includegraphics[width=0.9\linewidth]{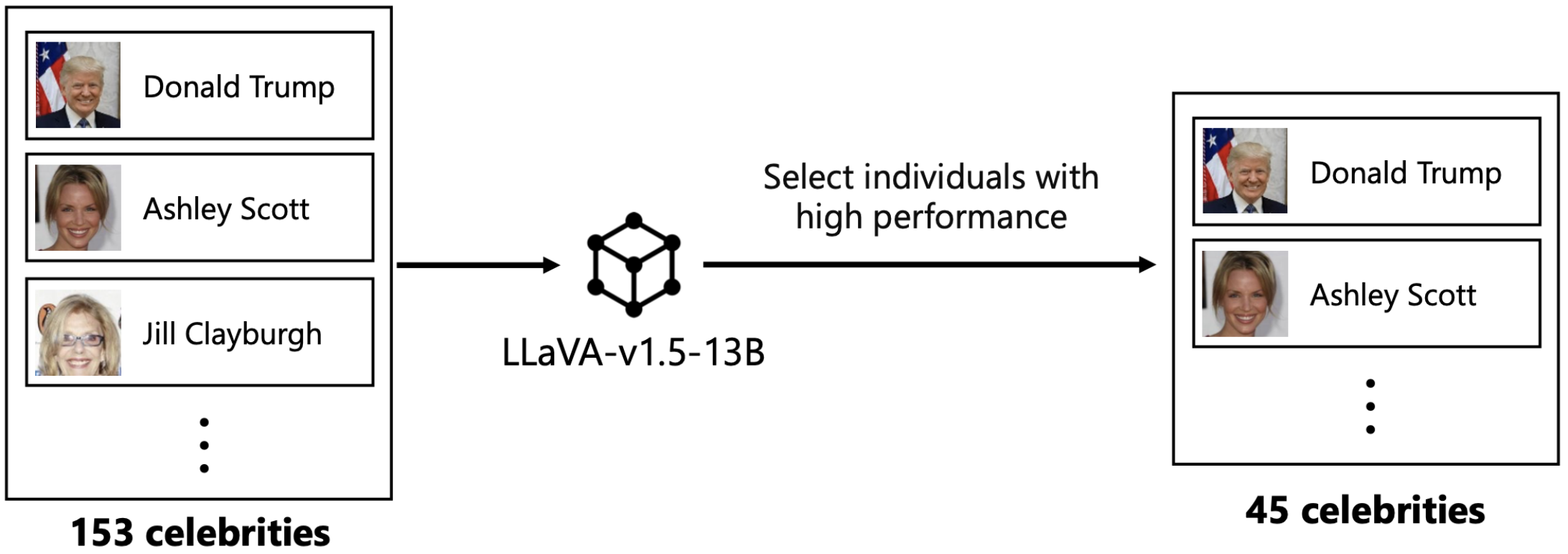}\\
   \caption{\textbf{Dataset Construction of Pre-trained Knowledge Unlearning.} 
We selected individuals with high performance on LLaVA and created dataset of pre-trained knowledge unlearning.}
    \label{fig:data_construction}
\end{figure}

\begin{table}[t]
    \centering
    \caption{
    \textbf{Comparison of Experimental Settings.} The data used in the experiments were selected from the publicly released MLLMU-Bench~\cite{mllmu_bench} dataset to match our experimental configurations.
    Each individual is associated with 10 questions—half text-only and half multimodal. Therefore, the dataset size equals the number of individuals multiplied by 10.
    }\label{tab:comparison}
    \setlength{\tabcolsep}{3pt}
    \resizebox{1\linewidth}{!}{
    \begin{tabular}{lcccc}
    \toprule
     & \makecell[c]{Type of Knowledge \\ to Unlearn}
     & \makecell[c]{Number of Individuals \\ in $\mathcal{D}_{\text{unlearn}}$}
     & Unlearning Count 
     & \makecell[c]{Individuals Unlearned \\ per Operation} \\
    \midrule
    \textbf{\makecell[c]{Fine-Tuned Knowledge Unlearning}}     & Fine-Tuning   & 50 & 1 & 50 \\
    \textbf{\makecell[c]{Pre-trained Knowledge Unlearning}}     & Pre-training   & 20 & 1 & 20 \\
    \textbf{Sustainability}                                    & Fine-Tuning   & 50 & 5 & 10 \\
    \bottomrule
    \end{tabular}
    }
\end{table}

\section{Experiments}

\subsection{Experimental Setup}

% Model, Unlearn method, dataset construction, dataset count 
In this study, we use LLaVA-v1.5-13B~\cite{llava} as the LMM. For experiment of fine-tuned knowledge unlearning and experiment of sustainability, we apply LoRA~\cite{lora} during both fine-tuning and unlearning. To evaluate unlearning, we use the accuracy on $\mathcal{D}_\text{unlearn}$ as the effectiveness metric, and the accuracy on $\mathcal{D}_\text{retain}$ together with the MMBench~\cite{mmbench} score as the generality metrics. MMBench is a standard benchmark for assessing an LMM's multimodal capabilities, such as object recognition and chart understanding.

\noindent\textbf{Unlearning Methods.}  
We evaluate the following unlearning techniques:  
(1) \textbf{Gradient Ascent (GA)}: Uses $\mathcal{D}_{\text{unlearn}}$ as unlearning data and updates parameters in the opposite direction of standard gradient descent to induce unlearning.
(2) \textbf{GA with KL Regularization (GA+KLR)~\cite{ga}}: To mitigate GA's tendency to degrade performance on retention tasks, adds a KL-divergence penalty to keep the updated model close to the original.
(3) \textbf{NPO}~\cite{npo}: A preference-tuning method that treats the unlearning data as negative examples without requiring positive examples.

\subsection{Dataset Construction}

\begin{figure}[t]
\centering
    \includegraphics[width=0.75\linewidth]{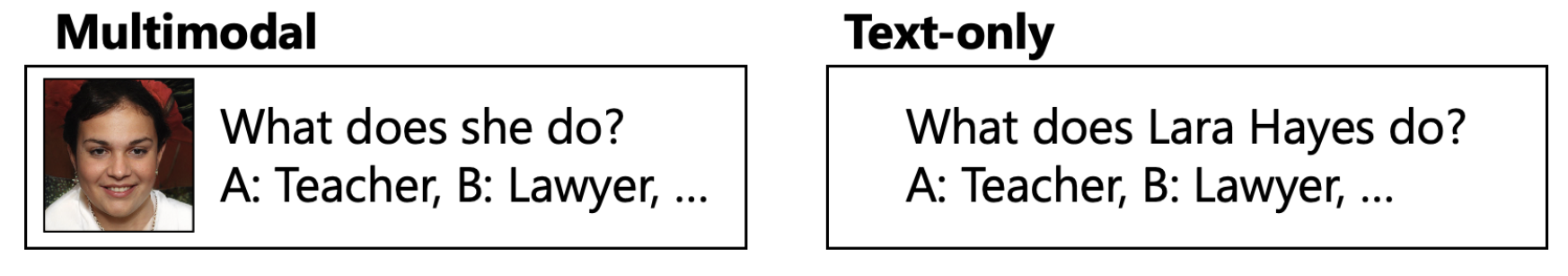}\\
   \caption{\textbf{Example of the Multimodal Task and the Text-only Task.} 
The multimodal task includes person\/'s face image, while the text-only task only has text prompt.}
    \label{fig:task}
\end{figure}

For our experiments, we use the dataset publicly released with MLLMU-Bench~\cite{mllmu_bench}. Each record contains one face image per fictional individual, along with ten question-answer pairs. Five of these pairs correspond to multimodal tasks, and the remaining five to text-only tasks. In both cases, the questions prompt personal details about the target individual (e.g., occupation, place of residence). The multimodal tasks include the person\/'s face image in the input, while the text-only tasks use only natural language input. Examples of these tasks are shown in Figure~\ref{fig:task}.
The experimental settings for each experiment of PULSE are shown in Table~\ref{tab:comparison}.

\noindent\textbf{Fine-tuned Knowledge Unlearning.}
LLaVA is fine-tuned on a dataset of 100 fictional individuals from MLLMU-Bench. Then 50 of those individuals are assigned to $\mathcal{D}_{\text{unlearn}}$ and the remaining 50 to $\mathcal{D}_{\text{retain}}$, after which unlearning is performed.

\noindent\textbf{Pre-trained Knowledge Unlearning.}
To evaluate how effectively pre-trained knowledge can be forgotten, the original model must already have a firm grasp of the target knowledge of unlearning. Consequently, as depicted in Figure~\ref{fig:data_construction}, we extracted only those celebrities from the MLLMU-Bench dataset for whom LLaVA-v1.5-13B attains high accuracy and used them in our experiments.
To ensure the pre-trained model initially possesses sufficient knowledge of the target individuals, we select 45 out of the 153 real famous individuals in the publicly released MLLMU-Bench dataset for which the LMM achieves high accuracy before unlearning. Of these, 20 are assigned to $\mathcal{D}_{\text{unlearn}}$ and 25 to $\mathcal{D}_{\text{retain}}$. We then perform unlearning on $\mathcal{D}_{\text{unlearn}}$ and evaluate the results. 
We note here that in the MLLMU-Bench, the subset of famous individuals is intended for assessing the model's generality after unlearning, whereas in our setting, it serves as part of the unlearning target itself.

\noindent\textbf{Sustainability.}
Figure~\ref{fig:teasers2} illustrates the pipeline for the sustainability experiment. In this experiment, the part of the  knowledge acquired through fine-tuning is designated as $\mathcal{D}_{\text{unlearn}}$ and is not unlearned in a single batch. Instead, $\mathcal{D}_{\text{unlearn}}$ is divided into five subsets, and unlearning is performed on a different individual in each run, for a total of five consecutive unlearning operations, whose performance we then evaluate.

\vskip\baselineskip

\begin{figure}[tb]
  \centering
  \begin{subfigure}[b]{0.48\linewidth}
    \centering
    \includegraphics[width=\linewidth]{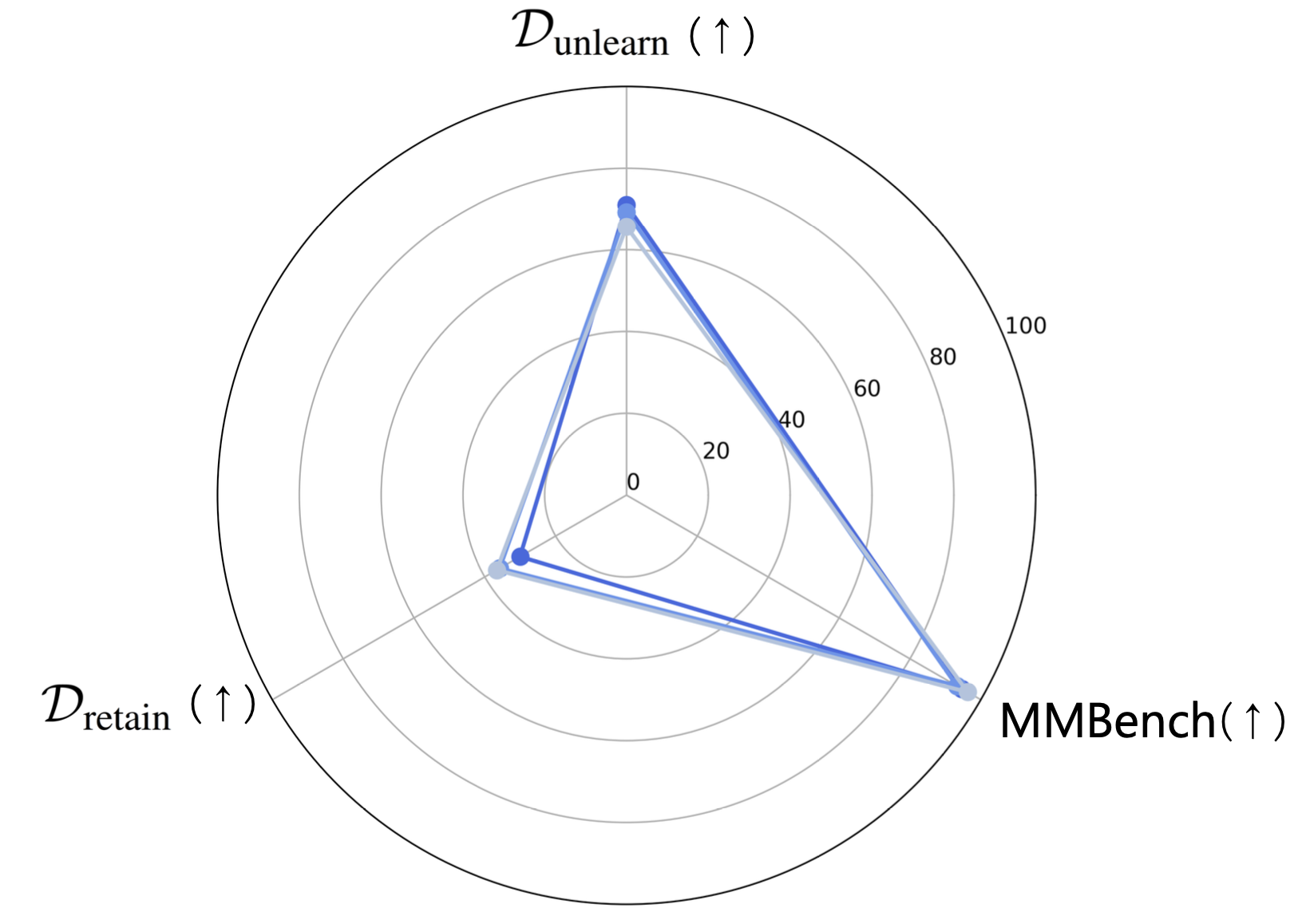}

    \includegraphics[width=0.75\linewidth]{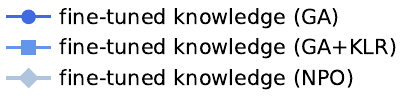}

    \captionsetup{labelformat=empty}
    \label{fig:raderA}
  \end{subfigure}
  \hfill
  \begin{subfigure}[b]{0.48\linewidth}
    \centering
    \includegraphics[width=\linewidth]{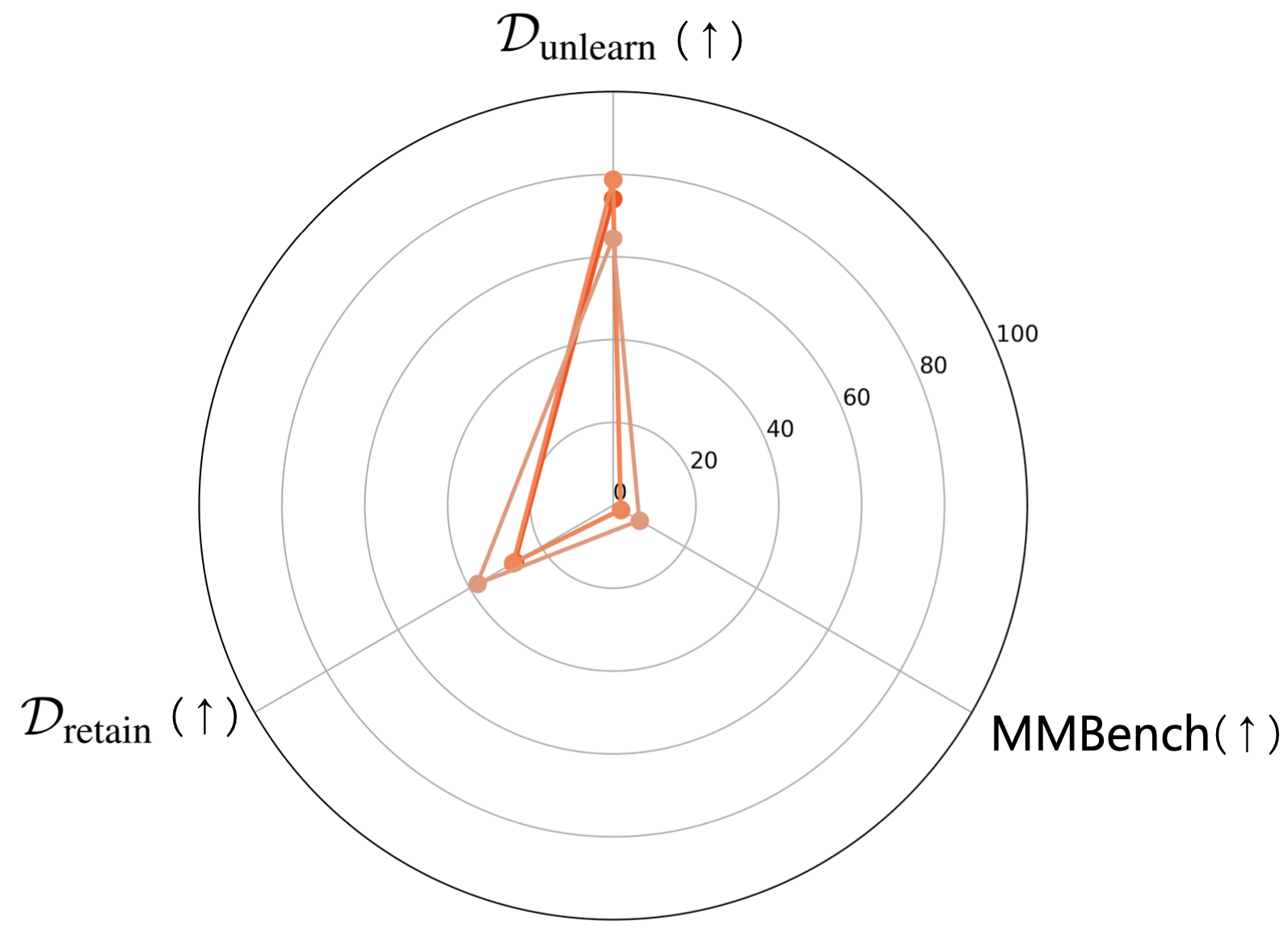}

    \includegraphics[width=0.75\linewidth]{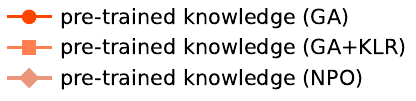}

    \captionsetup{labelformat=empty}
    \label{fig:raderB}
  \end{subfigure}

  \caption{\textbf{The Effect of the Source of Unlearning Target}.
  The $\mathcal{D}_{\text{unlearn}}$ axis shows what percentage of the model's pre-unlearning knowledge (set as 100) has been forgotten.
  For the $\mathcal{D}_{\text{retain}}$ and MMBench axes, it shows what percentage of pre-unlearning knowledge has been retained.
  All methods exhibit a substantial drop in MMBench score when unlearning pre-trained knowledge.}
  \label{fig:rader}
\end{figure}

\subsection{Main Results and Discussion}

\noindent\textbf{Unlearning Performance on Pre-trained Knowledge.}  
Figure~\ref{fig:rader} shows that, regardless of whether the unlearned knowledge was acquired through fine-tuning or through pre-training, accuracy on $\mathcal{D}_{\text{unlearn}}$ declines after unlearning, indicating that unlearning works to some degree in both setting. However, when we examine the MMBench accuracy, we find that unlearning fine-tuned knowledge reduces the original capability by at most about 10\%, whereas unlearning pre-trained knowledge leads to the loss of over 90\% of the original knowledge. This suggests (1) that pre-trained knowledge is harder to unlearn than fine-tuned knowledge, and (2) that this difficulty manifests as a substantial drop in post-unlearning generality. One possible explanation is that, during pre-training, the model learns relationships between the target individual and other entities, making it difficult to selectively unlearn only the target.

Notably, accuracy on $\mathcal{D}_{\text{retain}}$ also falls markedly. We attribute this to the domains of $\mathcal{D}_{\text{unlearn}}$ and $\mathcal{D}_{\text{retain}}$ being very similar, causing the model to unlearn both simultaneously. This finding is consistent with prior work~\cite{mllmu_bench}.

\begin{figure}[tb]
    \centering
    \begin{minipage}[c]{0.6\linewidth}
      \centering
      \includegraphics[width=\linewidth]{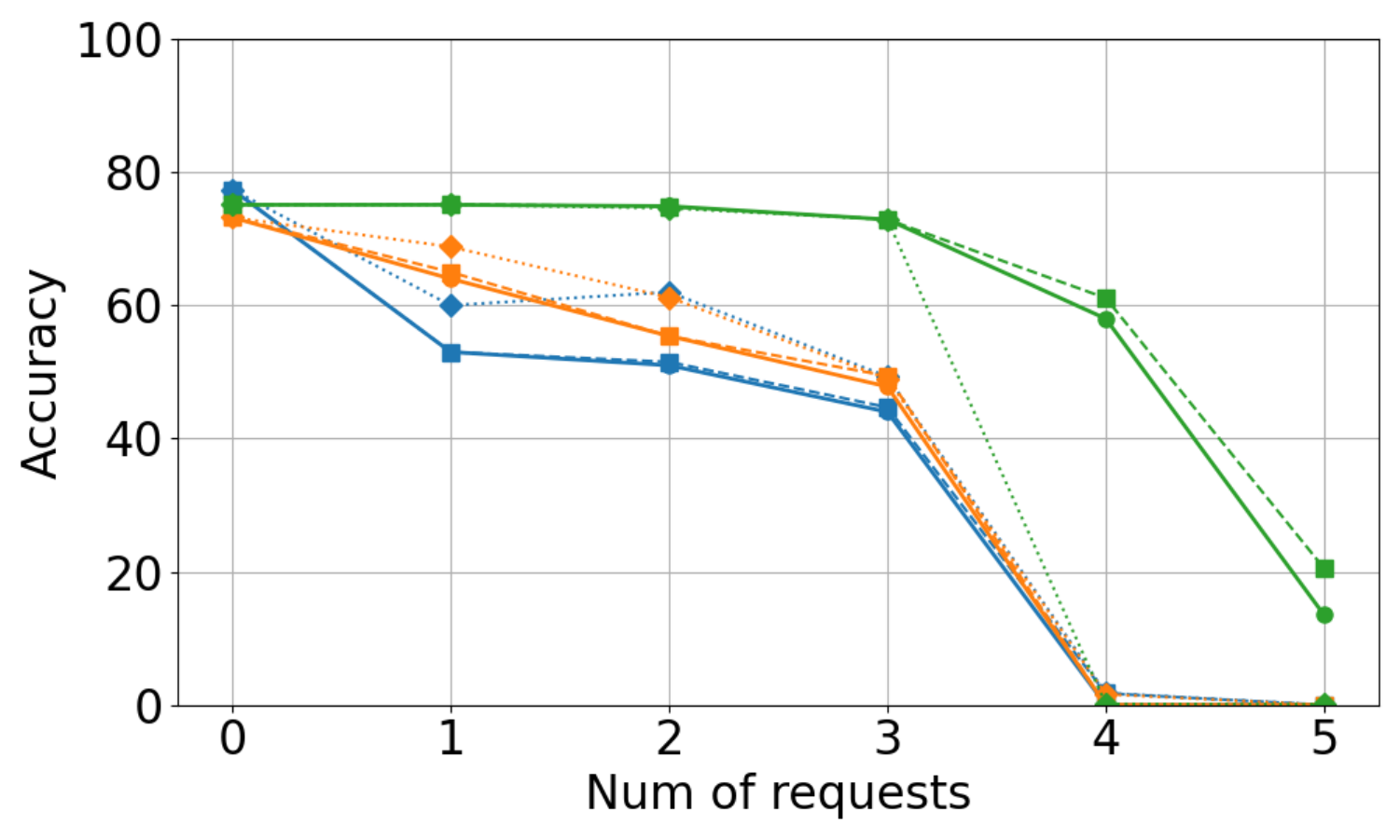}
    \end{minipage}%
    \hspace{0.05\linewidth}
    \begin{minipage}[c]{0.25\linewidth}
      \centering
      \includegraphics[width=\linewidth]{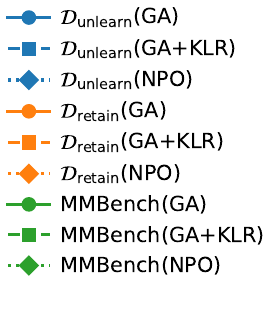}
    \end{minipage}
    \caption{\textbf{The Transition of Accuracy Over Multiple Requests.} All methods show a proper decrease in accuracy on $\mathcal{D}_{\text{unlearn}}$ as the number of unlearning requests increases, but at the same time accuracy on $\mathcal{D}_{\text{retain}}$ and MMBench also drops significantly, indicating that these methods fail to handle sequential requests sustainably.}
    \label{fig:sustainability_res}
\end{figure}

\noindent\textbf{Sustainability.}  
Figure~\ref{fig:sustainability_res} presents the results of the sustainability experiment. The horizontal axis denotes the number of unlearning operations applied consecutively to the same model. From these results, we observe that with repeated unlearning, not only does performance on $\mathcal{D}_{\text{unlearn}}$ degrade, but the generality metrics, which are accuracy on $\mathcal{D}_{\text{retain}}$ and MMBench, also gradually decline, such that after five unlearning operations, generality is almost completely lost. This finding reveals that current mainstream unlearning methods cannot maintain sustainability in LMM unlearning. We hypothesize that catastrophic forgetting occurs because repeated unlearning updates parameters that are also essential for retention tasks, leading to a rapid loss of previously acquired knowledge.  

\subsection{More Results and Discussion}

\noindent\textbf{Performance Differences by Task Modality.}  
In Table~\ref{tab:modal}, when the updated parameters include both the projection matrix and the language model (Proj, LLM), the accuracy on $\mathcal{D}_{\text{unlearn}}$ for ``Multi'' drops from 78.0\% to 9.6\%, whereas for ``Text'' it drops from 76.8\% to 35.2\%, indicating that the text-only task is more resistant to forgetting. One possible explanation is that including the projection matrix in the update target makes multimodal tasks easier to unlearn; however, even when updating only the LLM, ``Text'' still degrades less than ``Multi.'' Therefore, for a task such as querying the subject's place of residence (Figure~\ref{fig:task}), the model may fail on image-based queries but still succeed on text-only queries. Thus, applying existing unlearning methods to multimodal tasks may merely ``break the alignment between image and knowledge,'' casting doubt on whether the model has genuinely unlearned the target information.

Interestingly, we find that updating only the LLM significantly degrades performance on MMBench, whereas updating both the projection matrix and the LLM leads to only a slight drop. We hypothesize that allowing updates to the projection matrix makes it easier for the model to unlearn target samples by breaking the alignment between modalities. In contrast, restricting updates to the LLM alone makes the unlearning task harder and more disruptive to the model's general capabilities. A more rigorous investigation is left as an interesting avenue for future work.

\begin{table}[t]
    \centering
    \caption{\textbf{Performance Differences by Task Modality.} The ``Parameter Update Target'' column indicates which parts of LLaVA's parameters are updated during unlearning: ``Proj,LLM'' updates both the projection matrix between the image encoder and the language model (Proj) and the language model itself (LLM), while ``LLM'' updates only the language model. ``Multi'' denotes performance on multimodal tasks, and ``Text'' denotes performance on text-only tasks.}
    \label{tab:modal}
    \setlength{\tabcolsep}{3pt}
    \resizebox{0.75\linewidth}{!}{
    \begin{tabular}{ccccccc}
    \toprule
    \multirow{2}{*}{\shortstack{\textbf{Parameter}\\\textbf{Update Target}}} & \multirow{2}{*}{\textbf{Unlearning Method}} & \multicolumn{2}{c}{\textbf{$\mathcal{D}_\text{unlearn}$ ($\downarrow$)}} & 
    \multicolumn{2}{c}{\textbf{$\mathcal{D}_\text{retain}$ ($\uparrow$)}} &
    \multirow{2}{*}{\shortstack{\textbf{MMBench}\\\textbf{($\uparrow$)}}} \\
    & & Multi & Text & Multi & Text & \\
    \midrule
             & (Pre-unlearning) & 78.0 & 76.8 & 70.0 & 76.8 & 75.1 \\
    Proj,LLM & GA               &  9.6 & 35.2 & 14.8 & 29.2 & 71.1 \\
    LLM      & GA               & 24.8 & 33.2 & 29.2 & 34.4 & 48.8 \\
    \bottomrule
    \end{tabular}
    }
\end{table}

\section{Conclusion}

In this study, we proposed PULSE, a new evaluation protocol for unlearning in LMMs that addresses scenarios not covered by previous benchmarks. Our experiments revealed that, although unlearning knowledge acquired via fine-tuning in a single unlearning step can be moderately successful, existing methods such as GA, GA+KLR, and NPO suffer significant drops in model generality when applied to unlearning pre-trained knowledge or when repeated unlearning is required.

\bibliography{list.bib}

\begin{thebibliography}{15}
\providecommand{\natexlab}[1]{#1}
\providecommand{\url}[1]{\texttt{#1}}
\expandafter\ifx\csname urlstyle\endcsname\relax
  \providecommand{\doi}[1]{doi: #1}\else
  \providecommand{\doi}{doi: \begingroup \urlstyle{rm}\Url}\fi

\bibitem[Brown et~al.(2020)Brown, Mann, Ryder, Subbiah, Kaplan, Dhariwal,
  Neelakantan, Shyam, Sastry, Askell, et~al.]{gpt3}
Tom Brown, Benjamin Mann, Nick Ryder, Melanie Subbiah, Jared~D Kaplan, Prafulla
  Dhariwal, Arvind Neelakantan, Pranav Shyam, Girish Sastry, Amanda Askell,
  et~al.
\newblock Language models are few-shot learners.
\newblock \emph{NeurIPS}, 2020.

\bibitem[Yin et~al.(2023)Yin, Fu, Zhao, Li, Sun, Xu, and Chen]{lmm_survey}
Shukang Yin, Chaoyou Fu, Sirui Zhao, Ke~Li, Xing Sun, Tong Xu, and Enhong Chen.
\newblock A survey on multimodal large language models.
\newblock \emph{National Scienece Review}, 2023.

\bibitem[Jerez et~al.(2010)Jerez, Molina, Garc{\'\i}a-Laencina, Alba, Ribelles,
  Mart{\'\i}n, and Franco]{mu_for_linear}
Jos{\'e}~M Jerez, Ignacio Molina, Pedro~J Garc{\'\i}a-Laencina, Emilio Alba,
  Nuria Ribelles, Miguel Mart{\'\i}n, and Leonardo Franco.
\newblock Missing data imputation using statistical and machine learning
  methods in a real breast cancer problem.
\newblock \emph{AI in medicine}, 2010.

\bibitem[Schelter et~al.(2021)Schelter, Grafberger, and
  Dunning]{mu_for_ramdomtree}
Sebastian Schelter, Stefan Grafberger, and Ted Dunning.
\newblock Hedgecut: Maintaining randomised trees for low-latency machine
  unlearning.
\newblock In \emph{SIGMOD}, pages 1545--1557, 2021.

\bibitem[Yao et~al.(2024{\natexlab{a}})Yao, Xu, and Liu]{ga}
Yuanshun Yao, Xiaojun Xu, and Yang Liu.
\newblock Large language model unlearning.
\newblock \emph{NeurIPS}, 2024{\natexlab{a}}.

\bibitem[Eldan and Russinovich(2023)]{whos_hp}
Ronen Eldan and Mark Russinovich.
\newblock Who's harry potter? approximate unlearning in llms.
\newblock \emph{arXiv preprint arXiv:2310.02238}, 2023.

\bibitem[Li et~al.(2024)Li, Wei, Zhang, Qi, Du, Chen, and Bi]{siu}
Jiaqi Li, Qianshan Wei, Chuanyi Zhang, Guilin Qi, Miaozeng Du, Yongrui Chen,
  and Sheng Bi.
\newblock Single image unlearning: Efficient machine unlearning in multimodal
  large language models.
\newblock \emph{NeurIPS}, 2024.

\bibitem[Maini et~al.(2024)Maini, Feng, Schwarzschild, Lipton, and
  Kolter]{tofu}
Pratyush Maini, Zhili Feng, Avi Schwarzschild, Zachary~C Lipton, and J~Zico
  Kolter.
\newblock Tofu: A task of fictitious unlearning for llms.
\newblock \emph{COLM}, 2024.

\bibitem[Shi et~al.(2025)Shi, Lee, Huang, Malladi, Zhao, Holtzman, Liu,
  Zettlemoyer, Smith, and Zhang]{muse}
Weijia Shi, Jaechan Lee, Yangsibo Huang, Sadhika Malladi, Jieyu Zhao, Ari
  Holtzman, Daogao Liu, Luke Zettlemoyer, Noah~A Smith, and Chiyuan Zhang.
\newblock Muse: Machine unlearning six-way evaluation for language models.
\newblock \emph{ICLR}, 2025.

\bibitem[Liu et~al.(2025)Liu, Dou, Mengzhao~Jia, Zeng, Yuan, and
  Jiang]{mllmu_bench}
Zheyuan Liu, Guangyao Dou, Zhaoxuan~Tan Mengzhao~Jia, Qingkai Zeng, Yongle
  Yuan, and Meng Jiang.
\newblock Protecting privacy in multimodal large language models with
  mllmu-bench.
\newblock \emph{NAACL}, 2025.

\bibitem[Yao et~al.(2024{\natexlab{b}})Yao, Chien, Du, Niu, Wang, Cheng, and
  Yue]{yao2024machine}
Jin Yao, Eli Chien, Minxin Du, Xinyao Niu, Tianhao Wang, Zezhou Cheng, and
  Xiang Yue.
\newblock Machine unlearning of pre-trained large language models.
\newblock \emph{ACL}, 2024{\natexlab{b}}.

\bibitem[Zhang et~al.(2024)Zhang, Lin, Bai, and Mei]{npo}
Ruiqi Zhang, Licong Lin, Yu~Bai, and Song Mei.
\newblock Negative preference optimization: From catastrophic collapse to
  effective unlearning.
\newblock \emph{COLM}, 2024.

\bibitem[Liu et~al.(2024{\natexlab{a}})Liu, Li, Li, and Lee]{llava}
Haotian Liu, Chunyuan Li, Yuheng Li, and Yong~Jae Lee.
\newblock Improved baselines with visual instruction tuning.
\newblock \emph{CVPR}, 2024{\natexlab{a}}.

\bibitem[Hu et~al.(2022)Hu, Shen, Wallis, Allen-Zhu, Li, Wang, Wang, and
  Chen]{lora}
Edward~J Hu, Yelong Shen, Phillip Wallis, Zeyuan Allen-Zhu, Yuanzhi Li, Shean
  Wang, Lu~Wang, and Weizhu Chen.
\newblock Lora: Low-rank adaptation of large language models.
\newblock \emph{ICLR}, 2022.

\bibitem[Liu et~al.(2024{\natexlab{b}})Liu, Duan, Zhang, Li, Zhang, Zhao, Yuan,
  Wang, He, Liu, et~al.]{mmbench}
Yuan Liu, Haodong Duan, Yuanhan Zhang, Bo~Li, Songyang Zhang, Wangbo Zhao, Yike
  Yuan, Jiaqi Wang, Conghui He, Ziwei Liu, et~al.
\newblock Mmbench: Is your multi-modal model an all-around player?
\newblock In \emph{ECCV}, 2024{\natexlab{b}}.

\end{thebibliography}
\bibliographystyle{unsrtnat}
\end{document}